\documentclass[letterpaper]{article} % DO NOT CHANGE THIS
\usepackage{aaai24}  % DO NOT CHANGE THIS
\usepackage{times}  % DO NOT CHANGE THIS
\usepackage{helvet}  % DO NOT CHANGE THIS
\usepackage{courier}  % DO NOT CHANGE THIS
\usepackage[hyphens]{url}  % DO NOT CHANGE THIS
\usepackage{graphicx} % DO NOT CHANGE THIS
\usepackage{natbib}  % DO NOT CHANGE THIS AND DO NOT ADD ANY OPTIONS TO IT
\usepackage{caption} % DO NOT CHANGE THIS AND DO NOT ADD ANY OPTIONS TO IT
\usepackage{algorithm}
\usepackage{algorithmic}

\usepackage{amsmath}
\usepackage{amssymb}
\usepackage{multirow}

\usepackage{newfloat}
\usepackage{listings}
\DeclareCaptionStyle{ruled}{labelfont=normalfont,labelsep=colon,strut=off} % DO NOT CHANGE THIS
\floatstyle{ruled}
\newfloat{listing}{tb}{lst}{}
\floatname{listing}{Listing}
\title{LLM vs Small Model? Large Language Model Based Text Augmentation Enhanced Personality Detection Model}

\author {
    Linmei Hu\textsuperscript{\rm 1}\thanks{Corresponding author.},
    Hongyu He\textsuperscript{\rm 2},
    Duokang Wang\textsuperscript{\rm 2},
    Ziwang Zhao\textsuperscript{\rm 2},
    Yingxia Shao\textsuperscript{\rm 2},
    Liqiang Nie\textsuperscript{\rm 3}
}

\affiliations {
    \textsuperscript{\rm 1}School of Computer Science and Technology, Beijing Institute of Technology\\
    \textsuperscript{\rm 2}School of Computer Science, Beijing University of Posts and Telecommunications\\
    \textsuperscript{\rm 3}School of Computer Science and Technology, Harbin Institute of Technology (Shenzhen)\\
     hulinmei@bit.edu.cn, \{hehongyu, wangduokang, zhaoziwang, shaoyx\}@bupt.edu.cn, nieliqiang@gmail.com 
}

\usepackage{bibentry}
\begin{document}

\maketitle

\begin{abstract}
Personality detection aims to detect one's personality traits underlying in social media posts. One challenge of this task is the scarcity of ground-truth personality traits which are collected from self-report questionnaires. Most existing methods learn post features directly by fine-tuning the pre-trained language models under the supervision of limited personality labels. This leads to inferior quality of post features and consequently affects the performance. In addition, they treat personality traits as one-hot classification labels, overlooking the semantic information within them. In this paper, we propose a large language model (LLM) based text augmentation enhanced personality detection model, which distills the LLM's knowledge to enhance the small model for personality detection, even when the LLM fails in this task. Specifically, we enable LLM to generate post analyses (augmentations) from the aspects of semantic, sentiment, and linguistic, which are critical for personality detection. By using contrastive learning to pull them together in the embedding space, the post encoder can better capture the psycho-linguistic information within the post representations, thus improving personality detection. Furthermore, we utilize the LLM to enrich the information of personality labels for enhancing the detection performance. Experimental results on the benchmark datasets demonstrate that our model outperforms the state-of-the-art methods on personality detection.
\end{abstract}

\section{Introduction}

Personality is a combination of a person's internal characteristics, which can be reflected in their behavior \cite{survey}. Personality traits are defined by long-established personality theories such as the Myers-Briggs Type Indicator (MBTI) taxonomy (Myers-Briggs 1991). Personality detection aims to detect one's personality traits underlying  in social media posts. This emerging task in computational psycho-linguistics can provide specific personality information about the person and has been utilized in various applications, such as dialogue systems \citep{wen-etal-2021-automatically,yang-etal-2021-improving} and psychological treatments \citep{https://doi.org/10.1111/cpsp.12175}.

Traditional personality detection methods mostly rely on manually designed features, such as Linguistic Inquiry and Word Count (LIWC) \citep{LIWC}, that bring psychology theories and linguistic methods together. Recent works use end-to-end deep neural network to obtain text representations automatically in a data-driven manner \citep{data_driven_1,data_driven_2_bert,SN+attn,pandora}. However, extracting effective personality information from online posts is a non-trivial task. One main challenge is the scarcity of ground-truth personality traits, as they are collected from self-report questionnaires, which is time-consuming  and often raises concerns related to user privacy. To overcome the issue,  pre-trained language models have been applied to learn post representations \citep{TrigNet,attRCNN,TrigNet}.  To further improve the post representations, TrigNet \citep{TrigNet} considers the  contextual information of  a user's posts as well as the  external psycho-linguistic knowledge from LIWC \citep{attRCNN}. However, the learned post representations are still unsatisfactory, resulting in inferior performance.
In addition, they treat personality traits as one-hot classification labels, overlooking the semantic information within them. 

Recently, Large language models (LLMs) \citep{GPT3,ERNIE3,instuctGPT,OPT} have demonstrated significant capability in various natural language processing tasks under a generative format in zero-shot or few-shot scenarios. However, it  demonstrates unsatisfactory performance in personality detection compared to task-specific small model \cite{GPTPersonality}. Inferring people's personality traits from online posts is a complex and difficult task. 
Despite the ineffectiveness of LLMs in this task,  previous studies have demonstrated that LLMs exhibit strong language abilities, such as text comprehension, summarization, and sentiment analysis \cite{LLMsummarize,GPTSentiment,rational}, which can be used to distill useful knowledge for enhancing small models. On the one hand, we can leverage LLMs to generate post analyses (augmentations) from the aspects of semantic, sentiment, and linguistic, which are key factors for personality detection  \cite{On}. In this way, we can use the augmented post information to learn better post embeddings  for personality detection.
On the other hand, we can also use LLMs to generate explanations of the   complex personality labels to enrich the  label information for improving personality detection.

In this paper, we propose a large language model based text augmentation enhanced personality detection model, which distills useful knowledge from LLMs to  enhance a small model's personality detection capabilities, both on the data side and the label side. On the data side, we follow the methodology of contrastive sentence representation works \citep{simcse,llmcse}, using the LLM to generate \emph{knowledgeable} post augmentations from semantic, sentiment and linguistic aspects to provide personality-related knowledge. With the post augmentations, our contrastive post encoder can capture more psycho-linguistic information within the post representations. It is noteworthy that this method does not introduce extra costs during inference. On the label side, considering the personality labels are highly complex in their implications, which makes it difficult for detection, we also use the LLMs to generate additional explanations for the labels to enrich the label information for improving the detection performance.

In summary, our main contributions are as follows:
\begin{itemize}

    \item We propose a novel LLM based text augmentation enhanced personality detection model, which distills useful knowledge from LLMs to the small model, alleviating the issue of  data scarcity and improving personality detection. 
    \item Our model enables LLM to generate post augmentations for contrastive post representation learning from three specially designed aspects: semantic, sentiment, and linguistic, which are key factors for personality detection. Additionally, we use LLM to generate explanations of the personality labels for further improving the detection performance.
    \item  Experimental results on benchmark datasets have demonstrated that our model outperforms  the state-of-art baselines, which shows the effectiveness of our method.

\end{itemize}

\section{Related Work}

\subsection{Personality Detection}
Traditional personality detection relies on manual statistical feature engineering \citep{Tradi1,Tradi2}, such as extracting psycho-linguistic features from Linguistic Inquiry and Word Count (LIWC) \cite{LIWC} or statistical text features from the bag-of-words\cite{BagOfWord} model. As deep learning rapidly advances, a variety of Deep Neural Networks (DNNs) have been employed for personality detection tasks, resulting in significant success, including CNN \cite{attRCNN}, LSTM \citep{LSTM}, GRU \citep{SN+attn}, etc.
Recently, personality detection has benefited from pre-trained language models. \citet{data_driven_1} achieves promising performance by simply concatenating all the utterances from a single user into a document and encoding it with BERT \citep{bert} and RoBERTa \citep{roberta}. Some works improve it by leveraging the contextual information and external psycho-linguistic knowledge from LIWC  \citep{MD,TrigNet,zhuyangfu}. For example, Transformer-MD \citep{MD} stores posts' hidden states in the memory of Transformer-XL \citep{XL} in order to avoid introducing post-order bias. TrigNet \citep{TrigNet} constructs a heterogeneous graph between posts for each user based on the psycho-linguistic knowledge in LIWC and aggregates useful information with a  GAT.  D-DGCN \citep{DDGCN} builds a dynamic graph, which enables the model to learn the connections between the posts, and employs DGCN to integrate the information. Although these methods achieve promising performance, they still suffer from the limited supervision of insufficient personality labels, which leads to inferior quality of post embeddings that consequently affect the model performance. 

\subsection{Contrastive Sentence Representation Learning}

Contrastive learning was initially proposed by \citet{contras_1} and has been widely used for self-supervised representation learning in various domains. In the realm of NLP, a fundamental application of contrastive learning lies in sentence representation learning, applicable to both self-supervised and supervised scenarios. \citet{yan-etal-2021-consert} and \citet{simcse} propose different data augmentation strategies for contrastive learning using unlabeled data. In the supervised contrastive learning scenario, SimCSE \citep{simcse} demonstrated that supervised Natural Language Inference (NLI) datasets \citep{NLI_1,NLI_2} are effective for learning sentence embeddings by taking sentences with entailment labels as positive samples. Capitalizing on the strong capability of large language models, CLAIF \citep{CLAIF} utilizes LLM to generate similar text and similarity score, using them as positive samples and weights in info-NCE loss, thereby achieving improved performance. Different from these approaches, we generate data augmentations from the LLM, which contains task-specific information, to enhance personality detection performance.

\subsection{Knowledge Distillation from Large Language Models}

Knowledge distillation is used to transfer knowledge from a larger, more powerful teacher model, into smaller student models, enhancing their performance in practical applications \citep{hinton_distilling,distill_data_1,distill_data_2}. Recently, depending on the strong language abilities of LLMs, generating additional training data from LLMs for improving smaller models has become a new knowledge distillation trend \cite{self-instruct}. For example, Self-instruct \citep{self-instruct} distills instructional data to enhance the instruction-following capabilities of pre-trained language models. \citet{rational} generates rational text from LLMs to enhance the inference abilities of smaller models. Similarly, CLAIF \citep{CLAIF} leverages LLMs to produce similar texts, aiming to learn a better sentence representations. In this paper, we propose to leverage LLMs to generate post analyses (augmentations) from specially-designed perspectives to enhance the small model for personality detection.

\begin{figure}[t]
\centering
\includegraphics[width=1\columnwidth]{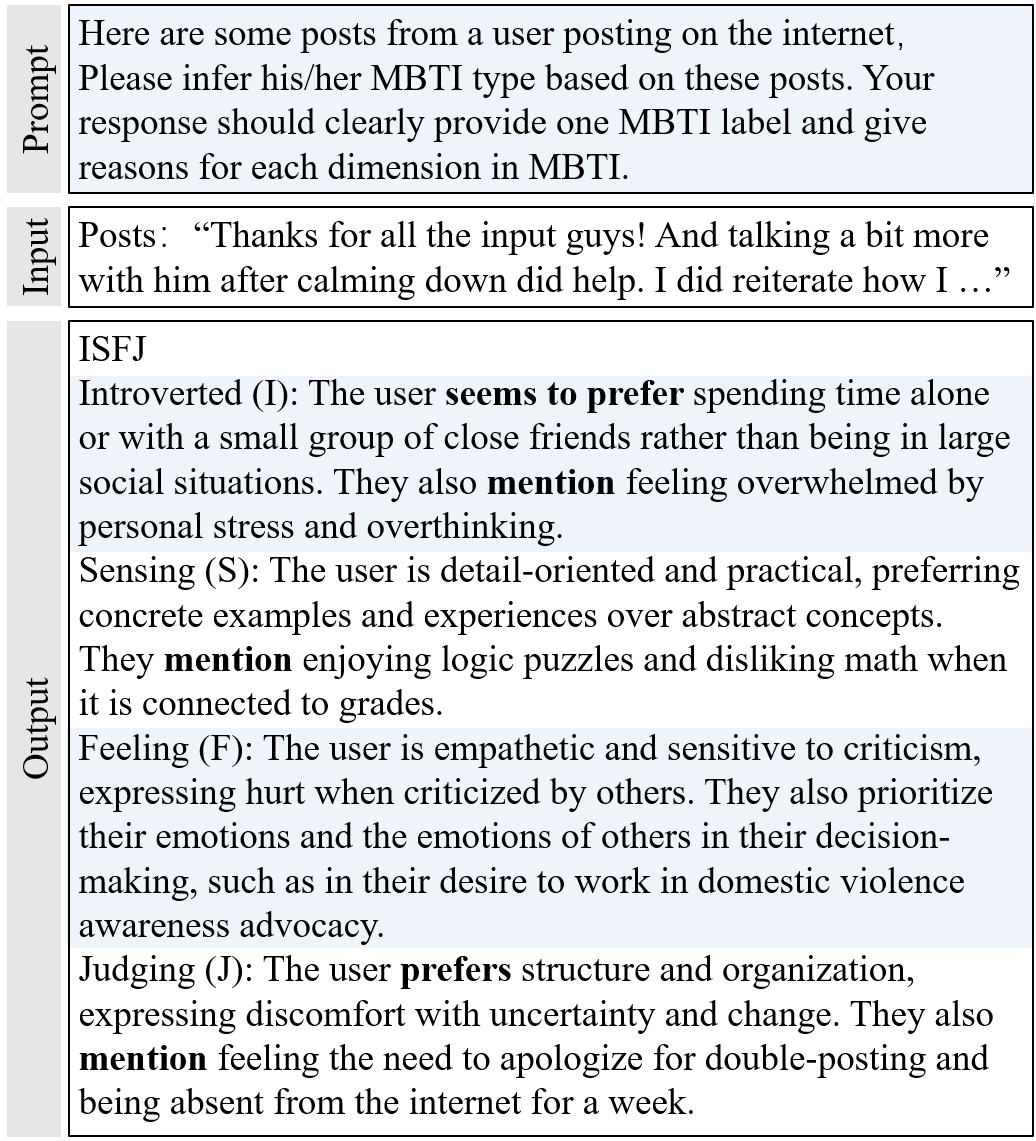} % Reduce the figure size so that it is slightly narrower than the column. Don't use precise values for figure width.This setup will avoid overfull boxes.
\caption{An example of personality detection using the LLM: The result provided by the LLM is ISFJ, while the actual ground truth is ENFP.}
\label{fig1}
\end{figure}

\section{Method}
Personality detection can be defined as a multi-document multi-label classification task \cite{TrigNet,DDGCN}. Formally, given a set $P=\{p_1, p_2 \ldots p_n\}$ of N posts from a user, where $p_i=\left[w_{i 1}, w_{i 2}, \ldots, w_{i m}\right]$ is $i$-th post with $m$ tokens. This task aims to  predict t-dimensional personality traits from the trait-specific label space $Y=\left\{\boldsymbol{y}_{1}, \boldsymbol{y}_{2}, \ldots, \boldsymbol{y}_{T}\right\}$ based for one user based on $P$. For MBTI taxonomy, $T = 4$ and $\boldsymbol{y}_t$ is one-hot vector. In this paper, we instruct the LLM to generate summarizations from semantical, sentimental and linguistic analysis for post augmentation. The augmented data represented as $X=\{P, P^\mathrm{s}, P^\mathrm{e}, P^\mathrm{l}\}$, where $P^\mathrm{s},P^\mathrm{e},P^\mathrm{l}$ correspond to the analysis texts for semantic, sentiment and linguistic, respectively. 
For the personality labels, we also utilize the LLM to generate explanations of the labels from semantic, sentiment and linguistic perspectives, thereby enriching the label information. For dimension $t$, the label descriptions are represented as $\hat{y}_t = \{L_{y_t,0},L_{y_t,1}\}$, where each $L_{y_t,j} = \{l_{y_t
,j}^\mathrm{s}, l_{y_t,j}^\mathrm{e}, l_{y_t,j}^\mathrm{l}\}$, corresponding to the semantic, sentiment and linguistic descriptions.

\subsection{Analysis of LLM Performance on Personality Detection}
LLMs show strong capabilities in many downstream tasks. However, their performance on the personality detection task is unsatisfying, as discussed in Section 4.6. Figure 1 shows a failure case of LLMs, where it wrongly classified ENFP as ISFJ. When we prompt the LLM to elucidate the analysis process underlying its classification, Figure 1 shows that the LLM primarily infers the personality traits based on only the post semantics. However, previous studies \cite{LIWC,survey} demonstrate that  the way people communicate and their sentiments often reveal more about their psychological state than the semantics of their communication content. LLM fails to capture the sentiment and linguistic patterns for personality detection.

\begin{figure*}[t]
\centering
\includegraphics[width=0.95\textwidth]{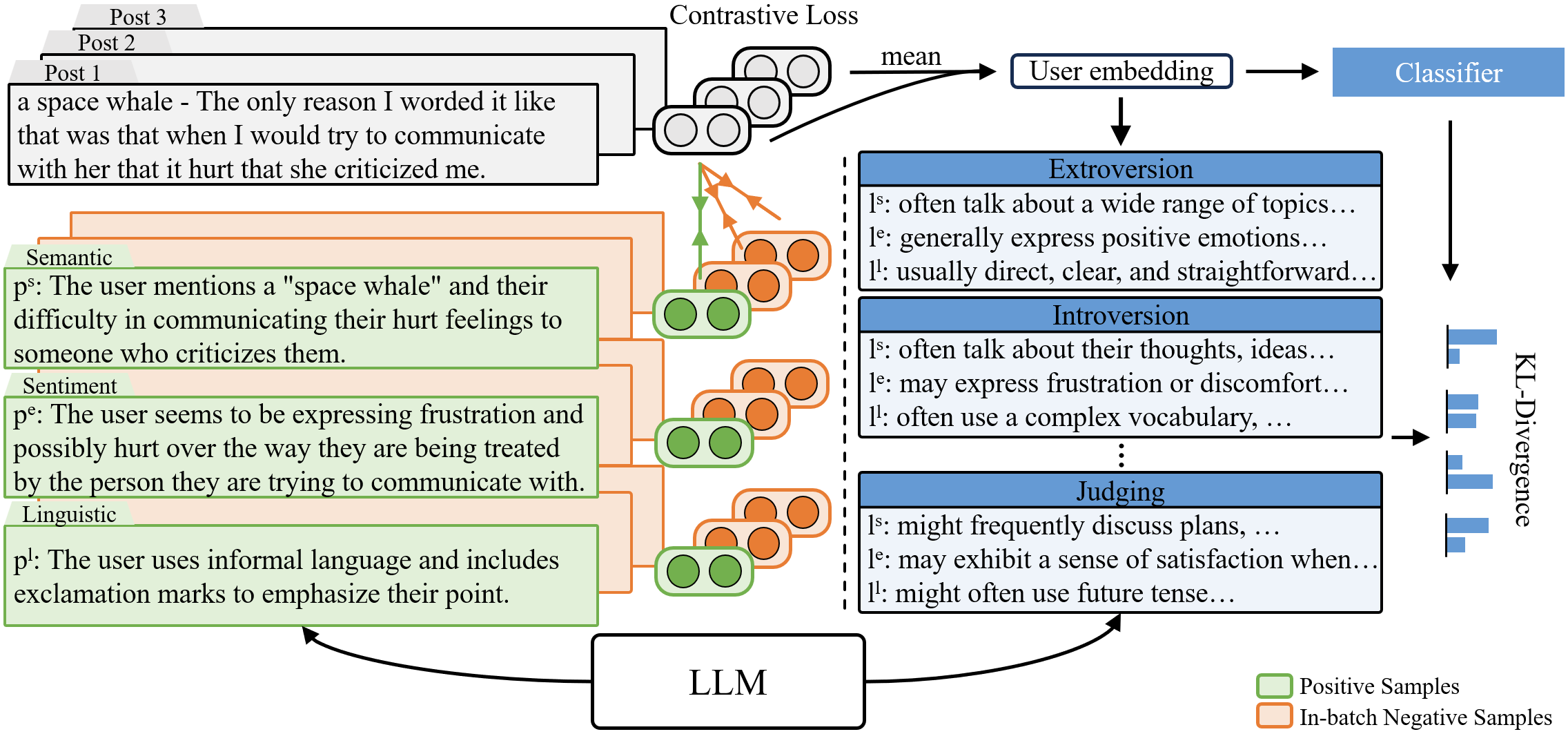} % Reduce the figure size so that it is slightly narrower than the column.
\caption{An overview of our TAE.}
\label{fig2}
\end{figure*}

\subsection{Generating Knowledgeable Post Augmentations from LLMs}

Personality detection is a complex and difficult task, since the personality traits are latent theoretical variables that cannot be directly or objectively observed \cite{On}.
%Despite of the ineffectiveness of the LLM in this task, previous studies have demonstrated that LLMs exhibit strong language abilities, such as text comprehension, summarization, and sentiment analysis \cite{LLMsummarize,GPTSentiment,rational}. 
Previous research has demonstrated a strong link between personality traits and a person's sentiments, words, and opinions \cite{personality_relate_3,emotional}. 
Therefore, despite of the ineffectiveness of the LLM in this task, we propose to fully utilize the LLM's abilities to distill personality-related knowledge, aiming to enhance smaller models for improving personality detection performance.

%As personality detection demonstrates a strong relationship with person's sentiments, words and opinions  \cite{personality_relate_3,emotional}, 
%we propose to take full use of the LLM's abilities to distill personality related knowledge to enhance the small models for improving personality detection performance. 

Specifically, we empirically instruct LLM (chatGPT) to generate post analyses from three aspects: semantic, sentiment, and linguistic, serving as data augmentations to original post. One analyses example is shown in Figure 2 generated with the instruction as follows:

\emph{Your task is to analyze the characteristics of a user based on a piece of text published by the user on the Internet. You are required to analyze it from the perspectives of semantic, sentiments, and linguistics. Note that if the text is incomplete and ends with an ellipsis, it may have been truncated due to external reasons, in which case you should ignore it. post:\ldots}

\subsection{Contrastive Post Encoder}

Contrastive learning aims to learn efficient representations by aligning semantically similar entities closer and distancing the dissimilar ones \citep{contras_1,simcse}. One key component of contrastive learning is the selection of positive pairs. 

With the obtained post augmentations from the LLM as positive samples, we learn the post embeddings with the contrastive post encoder that able to capture more personality-related feature.

Formally, for $i$-th post $p_i$ and it's analysis from three aspect $P_{pos} = \{p_i^{\mathrm{s}},p_i^{\mathrm{e}}, p_i^{\mathrm{l}}\}$,  we use the final hidden state of "[CLS]" as the sentence representation:
\begin{equation}
\boldsymbol{h_i} = \mathrm{BERT} \left(p_i\right) \in \mathbb{R}^{1 \times d},
\end{equation}
\begin{equation}
\boldsymbol{h^{+}_i} = \mathrm{BERT} \left(p^{+}_i\right), \text{where } p^{+}_i \in \{p^{\mathrm{s}}_i,p^{\mathrm{e}}_i, p^{\mathrm{l}}_i\}.
\end{equation}

Due to the distribution difference between the original post text and the analysis text, we add an extra MLP same to original BERT implementation as a projection head to mitigate this discrepancy:
\begin{equation}
  z_i= \delta \left( \boldsymbol{W}   \boldsymbol{h}_i + \boldsymbol{b} \right),
\end{equation}
\begin{equation}
  z^{+}_i= \delta \left( \boldsymbol{W}   \boldsymbol{h}_i^{+} + \boldsymbol{b} \right),
\end{equation}
where $\delta$ is Tanh function. We adopt the contrastive framework presented by Chen et al. (2020), utilizing a post-wise info-NCE loss with in-batch negatives, for $i$-th post in mini-batch:
\begin{equation}
\mathcal{L}_{cl} = \frac{1}{M} \sum_{i=1}^M \ell^{cl}_i
\end{equation}
\begin{equation}
\ell^{cl}_i  =-\log \sum_{(z_i,z_i^+)\in P_{pos}} \frac{e^{\operatorname{sim}\left(z_i, z_i^{+}\right) / \tau}}{\sum_{j=1}^M \sum_{(z_j,z_j^+)\in P_{pos}} e^{\operatorname{sim}\left(z_i, z_j^{+}\right) / \tau}}, \nonumber
\end{equation}
where M is total number of post in mini-batch, $\tau$ is temperature hyperparameter and $\operatorname{sim}\left(z_1, z_2\right)$ is cosine similarity $\frac{z_1^{\top} z_2}{\left\|z_1\right\| \cdot\left\|z_2\right\|}$. 

After obtaining posts representation of a user, we simply use average pooling method to produce the user representation:
\begin{equation}
\boldsymbol{u} = \mathrm{mean}\left ( \left [ \boldsymbol{h_1}, \boldsymbol{h_2}, \ldots, \boldsymbol{h_N}  \right ]  \right ).  
\end{equation}

\subsection{LLM Enriching Label Information}
Previous methods directly use one-hot label for personality detection, overlooking the information in personality traits. Since personality traits are complicated to comprehension, we also turn to  the LLM to generate explanations of each trait  from semantic, sentiment and linguistic perspectives, as the same to post augmentations. 

Specifically, given a personality label $L_{y_t,j}$, containing three descriptions from semantic, sentiment and linguistic aspects $L_{y_t,j} = \{l_{y_t,j}^\mathrm{s}, l_{y_t,j}^\mathrm{e}, l_{y_t,j}^\mathrm{l}\}$, the label representation is obtained by:
\begin{equation}
\boldsymbol{v_{y_t,j}} = \mathrm{mean} \left( \left[ \mathrm{BERT} \left(l_{y_t,j}\right) \right ] \right ), \text{where } l_{y_t,j} \in L_{y_t,j},
\end{equation}
where we also use the embedding of "[CLS]" in last layer as the sentence embedding.

Based on label representations, we generate soft labels to overcome the over-confident issue, which is primarily attributed to the information loss in the MBTI taxonomy's binary classification, which neglects the position on the scale \cite{survey}.  Dataset label noise introduced by the questionnaire measurement errors \cite{mea} could also lead to the over-confident issue. We leverage the generated label information to assign a soft label to a user based on the similarity between the user embedding and the label embedding. 

Formally, to generate soft label, we first calculate the dimension-wise label similarity distribution with cosine similarity and softmax function. Then, we combine the original one-hot vector with a controlling parameter $\alpha$ and apply an additional softmax function:
\begin{gather}
\boldsymbol{y}^s_t = \mathrm{softmax}\left(\mathrm{sim}\left ( \boldsymbol{u}, \boldsymbol{V_{y_t}} \right )\right) \label{eqn-5}, \\
\boldsymbol{y}^c_t = \mathrm{softmax}\left ( \alpha \boldsymbol{y}_t + \boldsymbol{y}^s_t \right ), \label{eqn-6}
\end{gather}

In this way, we can assign a softer label instead of one-hot label when a user is relatively neutral within a dimension, improving model's generalization ability.

Finally, we employ $T$ softmax-normalized linear transformations to predict personality traits, and apply a KL-divergence as the loss function:

\begin{equation}
\boldsymbol{\hat{y}}^t=\mathrm{softmax}\left(u\boldsymbol{W}_{\mathbf{u}}^t+\boldsymbol{b}_{\mathbf{u}}^t\right),
\end{equation}

\begin{equation}
\mathcal{L}_{det} = -\frac{1}{B} \sum_{i=1}^B \ell^{det}_i ,\nonumber
\end{equation}

\begin{equation}
\begin{aligned}
\ell^{det}_i &= \sum_{t=1}^T\text { KL-divergence }\left(\boldsymbol{y}^{c}, \boldsymbol{\hat{y}}^{t}\right) \\
&= \sum_{t=1}^T\sum_{j=0,1} \boldsymbol{y}^{c}_j \log \left(\frac{\boldsymbol{y}^{c}_j}{\boldsymbol{\hat{y}}^{t}_j}\right)
\end{aligned}
\end{equation}
where B denotes number of samples in batch.
Overall, the whole training objective of TAE is formulated as follows:
\begin{equation}
\mathcal{L} = \mathcal{L}_{det} + \lambda \mathcal{L}_{cl},
\end{equation}
where $\lambda$ is a trad-off parameter to balance the two losses. 

\begin{table}[t]
\centering
%\resizebox{.95\columnwidth}{!}{
\begin{tabular}{c|cccc}
\hline Dataset & Types & Train & Validation & Test \\
\hline \multirow{4}{*}{ Kaggle } & I / E & $4011 / 1194$ & $1326 / 409$ & $1339 / 396$ \\
& S / N & $727 / 4478$ & $222 / 1513$ & $248 / 1487$ \\
& T / F & $2410 / 2795$ & $791 / 944$ & $780 / 955$ \\
& P / J & $3096 / 2109$ & $1063 / 672$ & $1082 / 653$ \\

\hline \multirow{4}{*}{ Pandora } & I / E & $4278 / 1162$ & $1427 / 386$ & $1437 / 377$ \\
& S / N & $610 / 4830$ & $208 / 1605$ & $210 / 1604$ \\
& T / F & $3549 / 1891$ & $1120 / 693$ & $1182 / 632$ \\
& P / J & $3211 / 2229$ & $1043 / 770$ & $1056 / 758$ \\
\hline
\end{tabular}
\caption{Statistics of the Kaggle and Pandora datasets.}
\label{table1}
\end{table}

\section{Experiments}

\subsection{Datasets}
Following previous studies \citep{MD,TrigNet,DDGCN}, we choose two widely used datasets in personality detection, Kaggle\footnote{https://www.kaggle.com/datasnaek/mbti-type} and Pandora\footnote{https://psy.takelab.fer.hr/datasets/all} \citep{pandora}. The Kaggle dataset is collected from PersonalityCafe\footnote{http://personalitycafe.com/forum}, a platform where people share their personality types and engage in daily communications. This dataset contains a total of 8675 users, with each user contributing 45-50 posts. Pandora is another dataset collected from Reddit\footnote{https://www.reddit.com}, where personality labels are extracted from users' self-introductions that contain MBTI types. This dataset contains dozens to hundreds of posts from each of the 9067 users. Both datasets are based on the MBTI taxonomy, which divides people's personality into four dimensions, each containing two aspects: \textbf{I}ntroversion vs. \textbf{E}xtroversion (I vs. E), \textbf{S}ensing vs. i\textbf{N}tuition (S vs. N), \textbf{T}hinking vs. \textbf{F}eeling (T vs. F), and \textbf{P}erception vs. \textbf{J}udging (P vs. J). Since the two datasets are severely imbalanced, we employ the Macro-F1 metric, and use average Macro-F1 for overall performance. We adopt the same data division as in \cite{DDGCN}, which shuffles the datasets and splits them in a 60-20-20 proportion for training, validation, and testing, respectively. Table 1 shows the statistics of the two datasets. 

\begin{table*}[t]
\centering
\begin{tabular}{l|ccccc|ccccc}
\hline
\multirow{2}{*}{\textbf{Methods}} & \multicolumn{5}{c|}{\textbf{Kaggle}}                                                                                                  & \multicolumn{5}{c}{\textbf{Pandora}}                                                                                                  \\ \cline{2-11} 
                                  & \textit{\textbf{I/E}} & \textit{\textbf{S/N}} & \textit{\textbf{T/F}} & \multicolumn{1}{c|}{\textit{\textbf{P/J}}} & \textbf{Average} & \textit{\textbf{I/E}} & \textit{\textbf{S/N}} & \textit{\textbf{T/F}} & \multicolumn{1}{c|}{\textit{\textbf{P/J}}} & \textbf{Average} \\ \hline
SVM                               & 53.34                 & 47.75                 & 76.72                 & \multicolumn{1}{c|}{63.03}                 & 60.21            & 44.74                 & 46.92                 & 64.62                 & \multicolumn{1}{c|}{56.32}                 & 53.15            \\
XGBoost                           & 56.67                 & 52.85                 & 75.42                 & \multicolumn{1}{c|}{65.94}                 & 62.72            & 45.99                 & 48.93                 & 63.51                 & \multicolumn{1}{c|}{55.55}                 & 53.50            \\
BiLSTM                            & 57.82                 & 57.87                 & 69.97                 & \multicolumn{1}{c|}{57.01}                 & 60.67            & 48.01                 & 52.01                 & 63.48                 & \multicolumn{1}{c|}{56.21}                 & 54.93            \\
BERT\textsubscript{concat}        & 58.33                 & 53.88                 & 69.36                 & \multicolumn{1}{c|}{60.88}                 & 60.61            & 54.22                 & 49.15                 & 58.31                 & \multicolumn{1}{c|}{53.14}                 & 53.71            \\
BERT\textsubscript{mean}          & 64.65                 & 57.12                 & 77.95                 & \multicolumn{1}{c|}{65.25}                 & 66.24            & 56.60                 & 48.71                 & 64.70                 & \multicolumn{1}{c|}{56.07}                 & 56.52            \\
AttRCNN                           & 59.74                 & 64.08                 & 78.77                 & \multicolumn{1}{c|}{66.44}                 & 67.25            & 48.55                 & 56.19                 & 64.39                 & \multicolumn{1}{c|}{57.26}                 & 56.60            \\
SN+Attn                           & 65.43                 & 62.15                 & 78.05                 & \multicolumn{1}{c|}{63.92}                 & 67.39            & 56.98                 & 54.78                 & 60.95                 & \multicolumn{1}{c|}{54.81}                 & 56.88            \\
Transformer-MD                    & 66.08                 & \textbf{69.10}        & 79.19                 & \multicolumn{1}{c|}{67.50}                 & 70.47            & 55.26                 & 58.77                 & 69.26                 & \multicolumn{1}{c|}{\textbf{60.90}}        & 61.05            \\
TrigNet                           & 69.54                 & 67.17                 & 79.06                 & \multicolumn{1}{c|}{67.69}                 & 70.86            & 56.69                 & 55.57                 & 66.38                 & \multicolumn{1}{c|}{57.27}                 & 58.98            \\
D-DGCN                            & 68.41                 & 65.66                 & 79.56                 & \multicolumn{1}{c|}{67.22}                 & 70.21            & 61.55                 & 55.46                 & \textbf{71.07}        & \multicolumn{1}{c|}{59.96}                 & 62.01            \\
D-DGCN+$\ell_0$                   & 69.52                 & 67.19                 & 80.53                 & \multicolumn{1}{c|}{68.16}                 & 71.35            & 59.98                 & 55.52                 & 70.53                 & \multicolumn{1}{c|}{59.56}                 & 61.40            \\ \hline
ChatGPT                           & 65.86                 & 51.69                 & 78.60                 & \multicolumn{1}{c|}{63.93}                 & 66.89            & 55.52                 & 49.79                 & 71.25                 & \multicolumn{1}{c|}{60.51}                 & 59.27            \\ \hline
TAE(our)                          & \textbf{70.90}        & 66.21                 & \textbf{81.17}        & \multicolumn{1}{c|}{\textbf{70.20}}        & \textbf{72.07}   & \textbf{62.57}        & \textbf{61.01}        & 69.28                 & \multicolumn{1}{c|}{59.34}                 & \textbf{63.05}   \\ \hline
\end{tabular}
\caption{Overall results of our TAE and baseline models in Macro-F1 (\%) score.}
\label{table2}
\end{table*}

\subsection{Baselines}
We compare our model with several baselines as follows.

\textbf{SVM} \citep{SVM} and XGBoost \cite{XGboost}: These methods concatenate all the posts of a user into a document first, and then utilize SVM or XGBoost for classification based on features extracted using bag-of-words models.

\textbf{BiLSTM} \citep{LSTM}: Glove is employed for generating word embeddings. Subsequently, a Bi-directional LSTM is used to encode each post, with the averaged post representation serving as the user representation for personality detection.

\textbf{BERT\textsubscript{concat}} \citep{bertconcat}: It concatenates a user's posts into an extended document and then employs BERT to encode this composite text for user representation.

\textbf{BERT\textsubscript{mean}} \citep{data_driven_2_bert}:  It uses BERT to encode each post individually, extracting the CLS embedding for the post representation. Then, it employs mean pooling to derive the user representation.

\textbf{AttRCNN} \citep{attRCNN}: It adopts a hierarchical structure, in which a variant of Inception (Szegedy et al., 2017) is utilized to encode each post and a CNN-based aggregator is employed to obtain the user representation. Besides, it considers psycho-linguistic knowledge by concatenating the LIWC features with the user representation.

\textbf{SN+Attn} \citep{SN+attn}: It is also a hierarchical network that employs GRU and attention mechanisms to encode sequences from both the word-level and post-level for user representation, using a pre-trained 200-dimensional word2vec model (Mikolov et al., 2013) for word embeddings.

\textbf{Transformer-MD} \citep{MD}: Transformer-MD draws inspiration from the Transformer XL, first employing a low-level encoder to individually encode each post, storing the CLS embeddings in memory. Then, a high-level encoder aggregates information from all other posts to further encode the post.

\textbf{TrigNet} \citep{TrigNet}: TrigNet constructs a tri-graph for each user consisting of posts, words, and word categories, based on LIWC dictionary. It then employs a modified GAT to encode the graph and uses average pooling to obtain a user representation.

\textbf{D-DGCN} \citep{DDGCN}: DDGCN employs a dynamic graph to model a user's posts, allowing the model to learn the connections between posts autonomously, and then uses DGCN to encode the graph and obtains a user representation. It has a variant {\textbf{D-DGCN}+{$\ell_0$}} representing adding $\ell_0$ norm with Hard Concrete distribution.

\textbf{ChatGPT}\footnote{https://chat.openai.com/}: We applied the 'gpt-3.5-turbo-0301' version of ChatGPT, and set the temperature to 0, making the outputs mostly deterministic for the identical inputs.

\subsection{Implementation Details}
All the deep learning models are implemented in PyTorch, and the optimizer used is Adam (Kingma and Ba 2014). The learning rate for the pre-trained post encoder is set to 1e-5, and for other parameters is set to 1e-3. We employed 'bert-base-uncased' from BERT as our post encoder. The mini-batch size is set to 8. The temperature $\tau$ is set to 0.07 and trade-off parameter $\lambda$ is set to 1.The controlling parameter $\alpha$ is set to 4. Following previous works \cite{TrigNet,DDGCN,MD}, We limit each user to a maximum of 50 posts, and limit each post and analysis text in both datasets to a maximum length of 70 words. Additionally, we replace words that match any personality label with $\langle\mathrm{type}\rangle$ to avoid information leaks \cite{TrigNet}. 
For the LLM, in consideration of cost and efficiency, we use 'gpt-3.5-turbo-0301' model to generate post augmentations.

\subsection{Overall Results}

The overall results are presented in Table 2. We can find that the proposed TAE consistently outperforms all the baselines on Macro-F1 scores. {Compared to the best baselines D-DGCN (71.35\% on average Macro-F1) and D-DCGN+l0 (62.01\%) respectively on Kaggle dataset and Pandora dataset,} our model improves them by 1.01\% and 1.68\%, respectively. The result demonstrates the superiority of our model in personality detection. We believe the reasons are two-fold: (1) Our model TAE benefits from the post augmentations generated by the LLM, enabling the contrastive post encoder to extract information that is more conducive to personality detection. (2) The generated additional explanations of personality labels effectively help in accomplishing the detection task. {We can also find that compared to BERT\textsubscript{mean}, our model has a marked improvement, with respective gains  (5.83\% on the Kaggle dataset and 6.53\% on the Pandora dataset). This demonstrates that data augmentations from LLM in data-scarce situations is highly advantageous. Finally, we can observe that methods leveraging external psycho-linguistic knowledge from LIWC, such as AttRCNN and TrigNet achieve relatively good results, validating the effectiveness of introducing external knowledge. Furthermore, our model achieves the best performance, indicating that the generated post analyses and label explanations  from large language models  can provide  effective information for personality detection.}

\begin{table}[t]
\centering
\resizebox{1\columnwidth}{!}{
\begin{tabular}{l|ccccc}
\hline
\multirow{2}{*}{\textbf{Methods}} & \multicolumn{5}{c}{\textbf{Kaggle}}                                                                                                   \\ \cline{2-6} 
                                  & \textit{\textbf{I/E}} & \textit{\textbf{S/N}} & \textit{\textbf{T/F}} & \multicolumn{1}{c|}{\textit{\textbf{P/J}}} & \textbf{Average} \\ \hline
TAE\textsubscript{w/o semantic}   & 71.24                 & 66.34                 & 80.61                 & \multicolumn{1}{c|}{67.43}                 & 71.40            \\
TAE\textsubscript{w/o sentiment}  & 70.14                 & 65.29                 & 80.03                 & \multicolumn{1}{c|}{69.55}                 & 71.25            \\
TAE\textsubscript{w/o linguistic} & 70.34                 & 65.69                 & 78.89                 & \multicolumn{1}{c|}{69.71}                 & 71.07            \\ \hline
TAE\textsubscript{w/o label}      & 70.57                 & 65.89                 & 81.90                 & \multicolumn{1}{c|}{69.78}                 & 72.02            \\ \hline
TAE\textsubscript{w/o All}        & 64.65                 & 57.12                 & 77.95                 & \multicolumn{1}{c|}{65.25}                 & 66.24            \\ \hline
TAE\textsubscript{Concat}         & 66.60                 & 64.28                 & 78.19                 & \multicolumn{1}{c|}{62.55}                 & 67.91            \\
TAE\textsubscript{WS}             & 64.63                 & 62.62                 & 77.68                 & \multicolumn{1}{c|}{64.30}                 & 67.31            \\ \hline
TAE                               & 70.90                 & 66.21                 & 81.17                 & \multicolumn{1}{c|}{70.20}                 & 72.07            \\ \hline
\end{tabular}}
\caption{Results of ablation study on Macro-F1 on the Kaggle dataset.}
\label{table3}
\end{table}

\subsection{Ablation Study}
To verify the importance of each component in our TAE model, we conduct an ablation study on the Kaggle dataset. First, we analyze the contributions of the post augmentations from each aspect. {As we can see from Table \ref{table3}, among the three aspects}, linguistic augmentation proves to be the most important one in our method, as the average Macro-F1 score declines most largely when it is removed. Furthermore, the semantic information is the least influential augmentation. This indicates that the semantic information is relatively less important for personality detection, which is consistent with the observations in previous works \cite{TrigNet, survey}. When removing the LLM based label information enrichment, the performance of the model slightly decreases. The ablation studies demonstrate that our model benefits from  the LLM generated post augmentations from semantic, sentiment and linguistic aspects, as well as the LLM based label information enrichment.
% 实验结果出来的得引一些文献

{To further explore the effectiveness of using the LLM-generated analysis texts as data augmentations for contrastive post representation learning, we conducted ablation experiments. We compared this approach with the variants that directly use these LLM-based analysis texts (post augmentations) as additional input. TAE\textsubscript{Concat} first encodes post and analyses respectively and then concatenates their embeddings. TAE\textsubscript{WS} denotes using weighted sum as the pooling method of the post features and post augmentation features. TAE\textsubscript{w/o All} denotes removing all components in TAE which is an original BERT with mean pooling. Table \ref{table4} shows their performance on the Kaggle dataset, we can observe that models taking the post augmentations as extra input (TAE\textsubscript{Concat}, TAE\textsubscript{WS}) outperform the baseline TAE\textsubscript{w/o All}, but they are all inferior to our TAE which is a contrastive model. This demonstrates the effectiveness of data augmentation and the contrastive learning paradigm. Moreover, the models taking analysis texts as extra input requires the LLM to generate analysis texts during inference stage, which is costly. We also visualize the weights learned by TAE\textsubscript{WS}  in Figure 3. It shows that linguistic analysis is more important, which is consistent with our prevous observation. }

\subsection{LLM Performance}
To analyze LLM's performance in the personality detection task, we conducted a series of experiments using ChatGPT. We assessed LLM's capabilities under three settings: zero-shot, CoT, and few-shot, using the Kaggle testing set. For the few-shot setting, we randomly selected 3 examples from the training set. Specifically, we employed the 'gpt-3.5-turbo-0301' version but switched to 'gpt-3.5-turbo-16k-0613' for the few-shot setting, as the input length exceeded the limit. As shown in Table \ref{table4}, the performance of ChatGPT is comparable to that of a small model that has been fine-tuned (BERT\textsubscript{mean}) for a specific task on Macro-F1 metrics. Under the CoT setting and few-shot setting, the classification performance slightly decreased. This indicates that the LLM's reasoning ability fails in personality detection. Directly applying LLM to the personality detection is not appropriate. Thus, in this paper, we consider leveraging the LLM to enhance the small model for personality detection by distilling useful knowledge from LLM to the small model.

\begin{table}[t]
\centering
\resizebox{1\columnwidth}{!}{
\begin{tabular}{l|ccccc}
\hline
\multirow{2}{*}{\textbf{Methods}} & \multicolumn{5}{c}{\textbf{Kaggle}}                                                                              \\ \cline{2-6} 
                                  & \textit{\textbf{I/E}} & \textit{\textbf{S/N}} & \textit{\textbf{T/F}} & \textit{\textbf{P/J}} & \textbf{Average} \\ \hline
ChatGPT                           & 65.86                 & 51.69                 & 78.60                 & 63.93                 & 66.89            \\
ChatGPT\textsubscript{cot}        & 65.13                 & 60.35                 & 75.73                 & 59.30                 & 65.12            \\
ChatGPT\textsubscript{3 shot}     & 70.61                 & 58.35                 & 76.58                 & 65.43                 & 67.74            \\ \hline
TAE                               & 70.90                 & 66.21                 & 81.17                 & 70.20                 & 72.07            \\ \hline
\end{tabular}}
\caption{ChatGPT performances on kaggle testing set.}
\label{table4}
\end{table}

\subsection{Effect of Trade-Off Parameter}

For trade-off parameter $\lambda$ in TAE, we searched in $\{0.5,1 ,1.5 ,2, 2.5 ,3,3.5 ,4\}$. Figure 4 demonstrates how the model performance changes with the $\lambda$ on the validation sets of Kaggle and Pandora datasets. We can observe that the Macro-F1 value first grows and reaches the highest value at $\lambda$=1 while it begins to drop when  $\lambda$ is larger than 1. This may be because that initially with the value of $\lambda$ increases, TAE can benefit more from the contrastive learning with the LLM genrated post aumgmentations,  which helps to learn better post representations. However, if $\lambda$ is set too large ($>1$), {the contrastive signal out weight the detection loss.
Overall, $\lambda$=1 can reach best balance between detection loss and contrastive loss on both Kaggle and Pandora datasets.}

\begin{figure}[t]
\centering
\includegraphics[width=0.8\columnwidth]{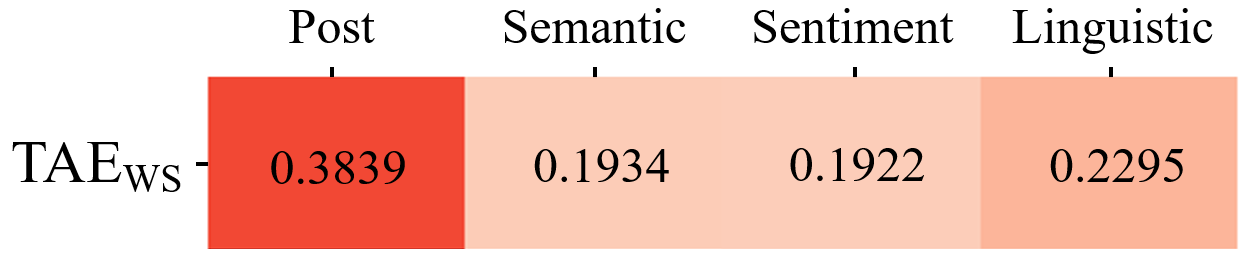} % Reduce the figure size so that it is slightly narrower than the column. Don't use precise values for figure width.This setup will avoid overfull boxes.
\caption{Visualization of learned weight of post and analyses in TAE\textsubscript{WS}.}
\label{fig3}
\end{figure}

\begin{figure}[t]
\centering
\includegraphics[width=0.8\columnwidth]{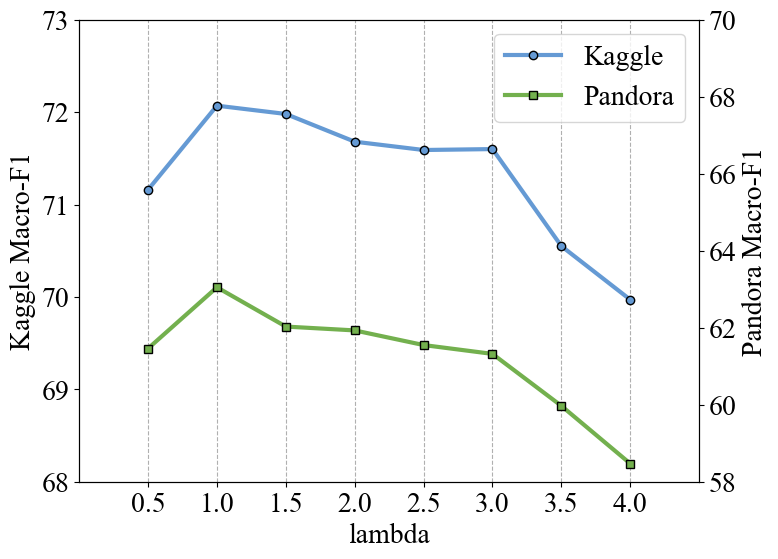} % Reduce the figure size so that it is slightly narrower than the column. Don't use precise values for figure width.This setup will avoid overfull boxes.
\caption{Performance curves for different trade-off parameter.}
\label{fig4}
\end{figure}

\section{Conclusion}

In this paper, we  propose a large language model (LLM) based text augmentation enhanced personality detection model, which distills the useful knowledge from the LLM to address the data scarcity issue faced by small models in personality detection, even when the LLM itself struggles with the task. By leveraging the LLM's abilities in text comprehension, summarization, and sentiment analysis, we instruct it to generate post analyses from three specially-designed perspectives: semantic, sentiment, and linguistic, which play a critical role for personality detection. Taking these analyses as positive samples and using contrastive learning to pull them together in the embedding space enables the post encoder in the small model to better capture the psycho-linguistic information within the post representations, thus improving personality detection. Furthermore, we utilize the LLM to generate label descriptions, enriching the semantics of personality labels and utilize label information to generate soft labels to overcome the over-confidence issue, enhancing models generalization ability.  Experimental results on two benchmark datasets demonstrate that our model outperforms the state-of-the-art methods on personality detection.

In future work, we will explore how to combine the advantages of the existing knowledge graphs and the LLM in improving personality detection.

\section{Acknowledgments}
This work was supported by the National Science Foundation of China (No. 62276029), and CCF-Zhipu. AI Large Model Fund (No. 202217).

\bibliography{aaai24}

\end{document}